\documentclass{article}

\PassOptionsToPackage{numbers, compress}{natbib}

\usepackage[preprint]{neurips_2022}

\usepackage[utf8]{inputenc} 
\usepackage[T1]{fontenc}    
\usepackage{hyperref}       
\usepackage{url}            
\usepackage{booktabs}       
\usepackage{amsfonts}       
\usepackage{amsmath}
\usepackage{nicefrac}       
\usepackage{microtype}      
\usepackage{xcolor}         
\usepackage{subcaption}
\usepackage{amsmath}
\usepackage{graphicx}
\usepackage{amsmath,amssymb,booktabs,tabulary,multirow,overpic,xcolor}
\usepackage{dsfont}

\usepackage{wrapfig}
\usepackage{verbatim}
\usepackage{float}
\floatstyle{plaintop}
\restylefloat{table}
\usepackage{graphicx}
\usepackage{amsmath}
\usepackage{amssymb}
\usepackage{booktabs}
\usepackage{soul}
\usepackage{paralist}
\usepackage[export]{adjustbox}

\DeclareMathOperator*{\argmin}{arg\,min}

\definecolor{pearThree}{HTML}{E74C3C}
\definecolor{pearDark}{HTML}{2980B9}
\definecolor{pearDarker}{HTML}{1D2DEC}

\hypersetup{
	colorlinks,
	citecolor=pearDark,
	linkcolor=pearThree,
    breaklinks=true,
	urlcolor=pearDarker}

\newcommand{\nameofmethod}{AvatarGen}

\newcommand{\eg}{\emph{e.g.}}
\newcommand{\ie}{\emph{i.e.}}

\title{Geometry-aware  Generative and Geometry-aware Adversarial Networks for Animatable Human Avatars}

\title{\nameofmethod{}: a 3D Generative    Model \\ for  Animatable Human Avatars}

\author{%
  Jianfeng Zhang$^{1}$\thanks{Equal contribution.} \ ,
  Zihang Jiang$^{1*}$,
  Dingdong Yang$^2$,
  Hongyi Xu$^2$,
  Yichun Shi$^2$,
  \\
  \textbf{Guoxian Song}$^2$,
  \textbf{Zhongcong Xu}$^1$,
  \textbf{Xinchao Wang}$^1$,
  \textbf{Jiashi Feng}$^2$
  \\
  $^1$National University of Singapore \quad
  $^2$ByteDance
}

\begin{document}

\maketitle

\begin{abstract}
Unsupervised generation of clothed virtual humans with various appearance and animatable poses is important for creating 3D human avatars and other AR/VR   applications. 
Existing   methods are  either  limited to   rigid object modeling, or not generative  and thus unable to 
synthesize high-quality virtual humans and animate them.
In this work, we propose \nameofmethod{}, the first method that enables not only non-rigid human generation with diverse appearance but also full control over poses and viewpoints,  while only requiring 2D images for training. 
Specifically, it extends the recent 3D GANs  to clothed human generation by utilizing
a coarse human body model as a proxy to warp the observation space into a standard avatar under a canonical space. 
To model non-rigid dynamics, it introduces a deformation network to learn pose-dependent deformations in the canonical space. 
To improve geometry quality of the generated human avatars, it leverages signed distance field as geometric representation, which allows more direct regularization from the body model on the geometry learning.
Benefiting from these designs, our method can generate animatable human avatars with high-quality appearance and geometry modeling, significantly outperforming previous 3D GANs.
Furthermore, it is competent for many applications, \eg, single-view reconstruction, reanimation, and text-guided synthesis.
Code and pre-trained model will be available.

\end{abstract}

\section{Introduction}

Generating  diverse and  high-quality virtual humans (avatars) with full control over their pose and viewpoint is a fundamental but extremely challenging task. Solving this task will benefit many applications like immersive photography visualization~\cite{zhang2022neuvv}, virtual  try-on~\cite{liu2021spatt}, VR/AR~\cite{xiang2021modeling,jiang2022selfrecon} and creative image editing~\cite{zhang2021editable,hong2022avatarclip}.

Conventional solutions rely on classical graphics modeling and rendering techniques~\cite{debevec2000acquiring,collet2015high,dou2016fusion4d,su2020robustfusion} to create avatars. Though offering high-quality, they typically require pre-captured templates, multi-camera systems, controlled studios, and long-term works of artists. In this work, we aim to make virtual human avatars widely accessible at low cost. Towards this goal, we propose the first 3D-aware avatar generative model  that can \emph{generate} 
\begin{inparaenum}[1)]
    \item high-quality virtual humans with 
    \item various  appearance styles, arbitrary poses and viewpoints,
    \item and be {trainable from only 2D  images, thus largely alleviating the effort to create virtual human.}
\end{inparaenum}

\begin{figure*}[t]
    \centering
    \small
    \includegraphics[width=0.95\linewidth]{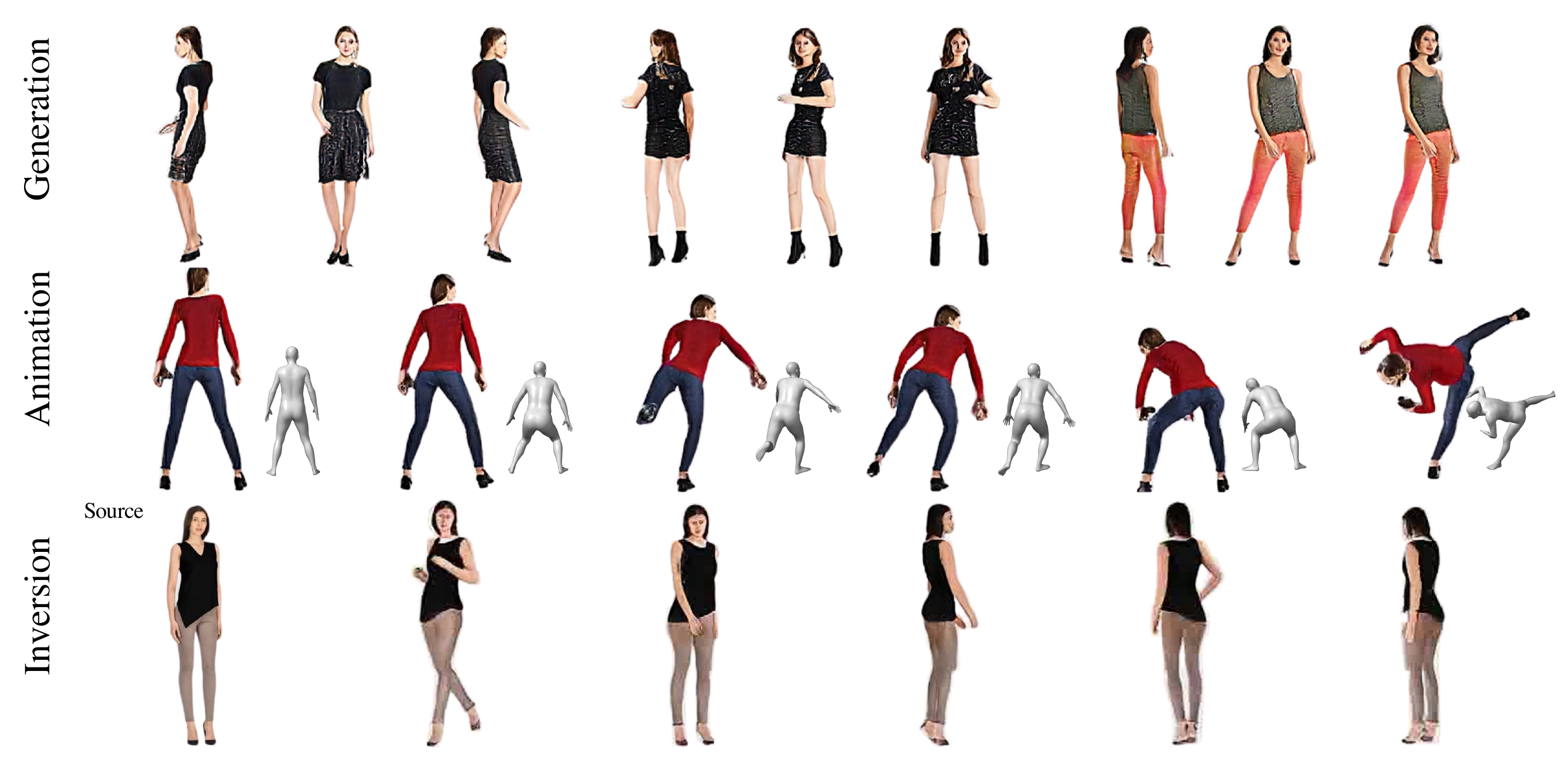}
    \caption{Our \nameofmethod{} model can  generate  clothed avatars with diverse appearance from arbitrary poses and viewpoints (top), animate the  avatars with specific pose signals (middle) and inverse  2D human images into animatable 3D avatars (bottom).} 
    \label{fig:teaser}
\end{figure*}

The 3D-aware generative models have recently seen rapid progress, fueled by introducing implicit neural representation (INR) methods~\cite{chen2019learning,Park_2019_CVPR,mescheder2019occupancy,mildenhall2020nerf} into generative adversarial networks \cite{chan2021pi,niemeyer2021giraffe,or2021stylesdf,gu2021stylenerf,chan2021efficient}.
However, these models are limited to  relatively simple and rigid objects, such as human faces and cars, and mostly fail to generate clothed human avatars whose appearance is highly sundry because of their articulated poses and great variability of clothing. Besides, they have limited control over the generation process and thus  cannot animate the generated objects, \ie, driving the objects to move by following certain instructions.
Another line of works leverage INRs~\cite{mildenhall2020nerf} to learn articulated human avatars for reconstructing a single subject from one's multi-view images or videos~\cite{peng2021animatable,2021narf,xu2021h,chen2021geometry,peng2022animatable}. While being able to animate  the avatars, these methods are \textit{not generative} and  cannot synthesize novel identities and appearances.

Aiming at    generative modeling of animatable human avatars, we propose \nameofmethod{}, the first model that can generate \emph{novel} human avatars with full  control over their poses and appearances. Our model is built upon  EG3D~\cite{chan2021efficient}, a recent   method that can generate  high quality   3D-aware {human faces}   via introducing  a new tri-plane representation  method.  However,     EG3D  is not directly applicable for clothed   avatar generation because it cannot handle the  challenges in modeling complex garments, texture, and   the articulated body structure with various  poses. Moreover, EG3D has limited control capability and thus it hardly animates the generated objects.

To address these challenges, we propose to decompose  the clothed avatar generation into  \textit{pose-guided canonical mapping} and \textit{canonical avatar generation}. Guided by a parametric human body model (\eg, SMPL~\cite{loper2015smpl}), our method warps  each  point in the observation space with a specified pose to a standard avatar with a fixed pre-defined pose in a canonical space via  an inverse-skinning transformation~\cite{huang2020arch}.
To accommodate  the non-rigid dynamics between the observation and canonical spaces (like clothes deformation), our method further trains a deformation module to predict the proper residual deformation.
As such, our method can generate arbitrary avatars in the observation space by deforming the canonical one which is much easier to generate and shareable  across different instances, thus largely alleviating the learning difficulties and achieving better appearance and geometry modeling.
Meanwhile, this formulation by design  enables \textit{disentanglement between the pose  and appearance}, offering   independent control over them.

Although the aforementioned method can generate  3D human avatars with reasonable geometry, we find it tends to produce noisy surfaces due to the lack of constraints on the learned geometry (density field). 
Inspired by recent works on neural implicit surface~\cite{wang2021neus,yariv2021volume,or2021stylesdf,peng2022animatable}, we propose to use a signed distanced field (SDF) {to impose stronger  \textit{geometry-aware guidance} for the model training}. 
Compared with the density field, SDF gives a better-defined surface representation, which facilitates more direct regularization on learning the avatar geometries. 
{Moreover, the model can leverage the coarse body model from SMPL to infer  reasonable signed distance values, which greatly improves quality of the clothed avatar generation and animation. }
The SDF-based volume rendering techniques~\cite{wang2021neus,yariv2021volume,or2021stylesdf} are used to render the low resolution feature maps, which are further decoded to high-resolution images with the StyleGAN generator~\cite{Karras2020stylegan2,chan2021efficient}. 

As shown in Fig.~\ref{fig:teaser}, trained from  2D images without using any multi-view or temporal information and 3D geometry annotations,   \nameofmethod{}  can generate a large variety of  clothed human with diverse appearances under arbitrary poses and viewpoints. 
We evaluate it quantitatively, qualitatively, and through a perceptual study; it strongly outperforms previous state-of-the-art methods.
Moreover, we demonstrate it on several applications, like single-view 3D reconstruction and text-guided synthesis.

Our contributions are threefold. 1) To our best knowledge,  \nameofmethod{} is the first model able to generate a large variety of animatable clothed human avatars without requiring multi-view, temporal or 3D annotated data.
2) We propose a human generation pipeline that achieves accurate appearance and geometry modeling, with full control over the pose and appearance. 
3) We demonstrate state-of-the-art 3D-aware human image synthesis on several benchmarks along with high-quality geometry.

\section{Related Works}

{\bf Generative 3D-aware image synthesis.}
Generative adversarial networks (GANs)~\cite{goodfellow2014generative} have recently achieved photo-realistic image quality for 2D image synthesis~\cite{karras2018progressive,karras2019style,Karras2020stylegan2,Karras2021}. Extending these capabilities to 3D settings has started to gain attention.
Early methods combine GANs with voxel~\cite{wu2016learning,hologan,nguyen2020blockgan}, mesh~\cite{Szabo:2019,Liao2020CVPR} or point cloud~\cite{achlioptas2018learning,li2019pu} representations for 3D-aware image synthesis.
Recently, several methods represent 3D objects by learning an implicit neural representation (INR)~\cite{schwarz2020graf,chan2021pi,niemeyer2021giraffe,chan2021efficient,or2021stylesdf,gu2021stylenerf,deng2021gram}. 
Among them, some methods use INR-based model as generator~\cite{schwarz2020graf,chan2021pi,deng2021gram}, while some others combine INR generator with 2D decoder for higher-resolution image generation~\cite{niemeyer2021giraffe,gu2021stylenerf,xue2022giraffe}. 
Follow-up works like EG3D~\cite{chan2021efficient} proposes an efficient tri-plane representation to model 3D objects, StyleSDF~\cite{or2021stylesdf} replaces density field with SDF for better geometry modeling and Disentangled3D~\cite{tewari2022d3d} represents objects with a canonical volume along with deformations to disentangle geometry and appearance modeling.
However, such methods are typically not easily extended to non-rigid clothed humans due to the complex pose and texture variations. 
Moreover, they have limited control over the generation process, making the generated objects hardly be animated. 
Differently, we study the problem of 3D implicit generative modeling of clothed human, allowing free control over the poses and appearances.

{\bf 3D human reconstruction and animation.}
Traditional human reconstruction methods require complicated hardware that is expensive for daily use, such as depth sensors~\cite{collet2015high,dou2016fusion4d,su2020robustfusion} or dense camera arrays~\cite{debevec2000acquiring,guo2019relightables}. 
To reduce the requirement on the capture device, some methods train networks to reconstruct human models from RGB images with differentiable renderers~\cite{xu2021texformer,Gomes2022}. 
Recently, neural radiance fields~\cite{mildenhall2020nerf} employ the volume rendering to learn density and color fields from dense camera views. Some methods augment neural radiance fields with human shape priors to enable 3D human reconstruction from sparse multi-view data~\cite{peng2021neural,chen2021geometry,xu2021h,su2021anerf}.
Follow-up improvements~\cite{peng2021animatable,chen2021animatable,liu2021neural,peng2022animatable,weng2022humannerf} are made by combining implicit representation with the SMPL model and exploiting the linear blend skinning techniques to learn animatable 3D human modeling from temporal data. 
However, these methods are not generative, \ie, they cannot synthesize novel identities and appearances. 
In this work, we learn fully generative modeling of human avatars from only 2D images, largely alleviating the cost to create virtual humans.

\begin{figure*}[t]
    \centering
    \small
    \includegraphics[width=1.0\linewidth]{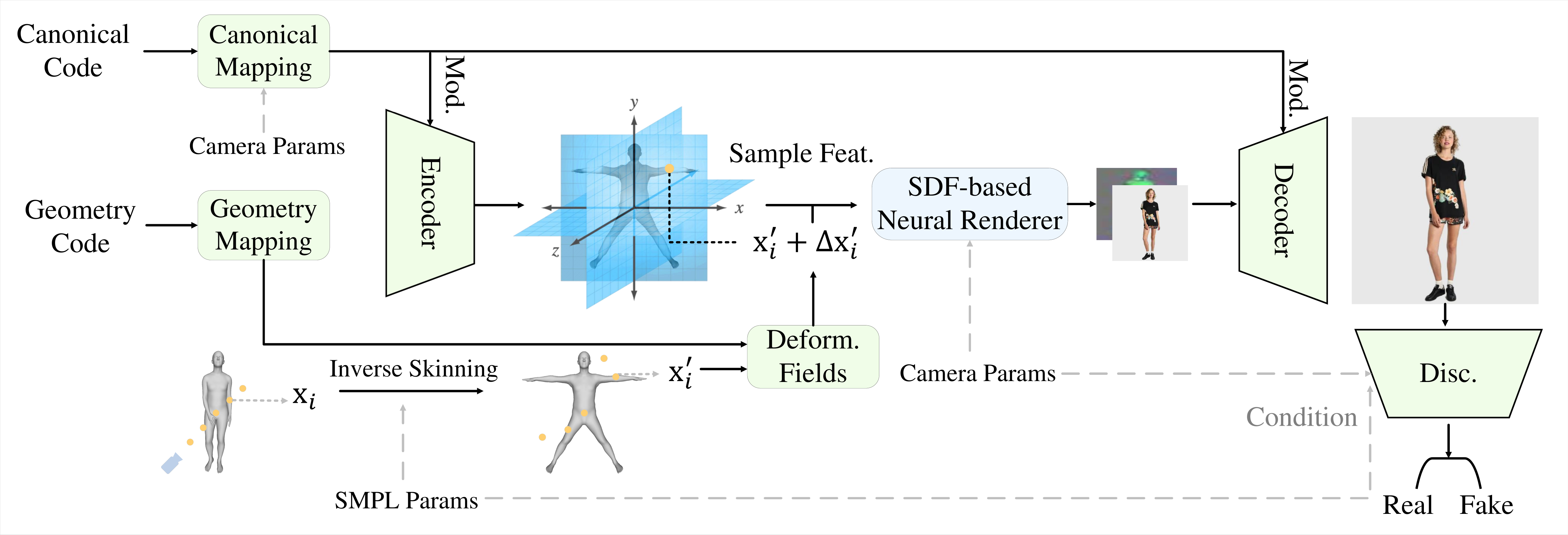}
    \caption{Pipeline of \nameofmethod{}. Taking the canonical code and camera parameters as input, the encoder   generates   tri-plane based features of a canonical posed human avatar. The geometry code is applied to modulate the deformation field module which   deforms the sampled points in the observation space to the canonical space under the guidance of the pose condition (SMPL parameters). The deformed spatial positions are   used to sample features on the tri-plane, which are then rendered as low-resolution  features and images using the SDF based neural renderer. Finally, the decoder   decodes the feature images to high resolution images. The generator  with is optimized a camera and pose conditioned discriminator via adversarial training. }
    \label{fig:arch}
\end{figure*}

\section{Method}

{Our goal is to build a generative model for diverse clothed 3D human avatar generation with varying  appearances in arbitrary poses. The model is trained from  2D images without using multi-view or temporal information and 3D scan annotations. Its framework is summarized in Fig.~\ref{fig:arch}.}

\subsection{Overview}

{\bf Problem formulation.}
We aim to train a 3D generative model $G$ for geometry-aware human  synthesis. Following EG3D~\cite{chan2021efficient}, we associate each training image with a set of camera parameters $\mathbf{c}$ and pose parameters $\mathbf{p}$ (in SMPL format~\cite{loper2015smpl}), which are obtained from  an off-the-shelf pose estimator~\cite{kolotouros2019spin}. Given a random latent code $\mathbf{z}$ sampled from Gaussian distribution, and a  new camera $\mathbf{c}$ and pose $\mathbf{p}$ as conditions, the generator $G$ can synthesize  a corresponding   human image  $I=G(\mathbf{z}|\mathbf{c},\mathbf{p})$. We optimize $G$  with a discriminator $D$ via adversarial training.

{\bf Framework.}
Fig.~\ref{fig:arch} illustrates the framework of our proposed generative model. It takes   camera parameters and SMPL parameters (specifying the generated pose) as inputs and generates  3D human avatar and its 2D images accordingly.  Our model jointly processes the random canonical code and camera parameters by a canonical mapping module (implemented by an MLP) to generate the intermediate latent code that modulates  the convolution kernels of the  encoder. This encoder then generates a tri-plane based features~\cite{chan2021efficient} corresponding to a canonical pose representation. Regarding the pose control, our model first jointly processes a random geometry code and the input pose condition via an MLP-based geometry mapping module, and outputs latent features. Then, {given a spatial point $\mathbf{x}$ in the observation space, the output latent features are processed by a deformation filed module, generating  non-rigid residual deformation $\Delta \mathbf{x}'$ over the inverse skinned point $\mathbf{x'}$ in the canonical space.}
We sample the features from the tri-plane according to the deformed spatial position $\mathbf{x'} + \Delta \mathbf{x'}$, which are then transformed into appearance prediction (i.e., color features) and geometry prediction (i.e., SDF-based features) for volume rendering. We will explain these steps in details in the following sections.

\subsection{Representations of 3D Avatars}
\label{sec:triplane}
It is important to choose an efficient approach to represent 3D human avatars. The recent  EG3D~\cite{chan2021efficient} introduces a memory-efficient tri-plane representation for 3D face modeling. It explicitly stores features on three axis-aligned orthogonal planes (called tri-planes), each of which  corresponds to a dimension within the 3D space. {Thus, the intermediate features of any   point  of a 3D face can be obtained  via simple lookup over the tri-planes, making the feature extraction  much more efficient than NeRF that  needs to forward  all the sampled 3D points through MLPs~\cite{mildenhall2020nerf}.}
Besides, the tri-plane representation effectively decouples the feature generation  from volume rendering, and can be directly generated from more efficient CNNs instead of MLPs. 

Considering  these benefits, we also choose   the tri-plane representation for 3D avatar modeling. 
{However, we found  directly adopting it for clothed human avatars generation  results in poor quality. Due to the much higher degrees of freedom of human bodies than faces,  it is very challenging  for the naive tri-plane representation model to learn pose-dependent appearance and geometry from only 2D images.} We  thus propose the following new approaches to address the difficulties.

\subsection{Generative 3D Human  Modeling}
\label{sec:decompose}

There are two main challenges for 3D human  generation. The first   is how to effectively integrate  pose condition into the tri-plane representations, making the generated human pose fully controllable and animatable.  One naive way is to combine the pose condition $\mathbf{p}$ with the latent codes  $\mathbf{z}$ and $\mathbf{c}$ directly, and feed them to the encoder to generate   tri-plane features. However,  such naive design cannot achieve high-quality synthesis and animation due to limited pose diversity and insufficient geometry supervision. Besides, the pose and appearance   are tightly entangled, making independent control impossible. The second challenge is that learning pose-dependent clothed human appearance and geometry from 2D images only is highly under-constrained, making the model training  difficult and    generation quality poor. 

To tackle these challenges, our \nameofmethod{} decomposes the  avatar generation    into two steps: \textit{pose-guided canonical mapping} and \textit{canonical avatar generation}.  Specifically, \nameofmethod{} uses  SMPL model~\cite{loper2015smpl} to parameterize the underlying 3D human body.  With a pose parameterization,  \nameofmethod{} can easily deform a 3D point $\mathbf{x}$ within the observation space with pose   $\mathbf{p}_o$ to a canonical pose $\mathbf{p}_c$ (an ``X''-pose as shown in Fig.~\ref{fig:arch}) via Linear Blend Skinning~\cite{jacobson2012fast}. 
Then, \nameofmethod{} learns to generate the appearance and geometry of human avatar in the canonical space. The canonical space is shared across different instances with a fixed template pose $\mathbf{p}_c$ while its appearance and geometric details can be varied  according to the latent code $\mathbf{z}$, leading to generative human modeling. 

Such a task factorization  scheme facilitates  learning of a generative canonical  human avatars  and effectively helps the model generalize to unseen poses, achieving animatable clothed human avatars generation. Moreover, it by design disentangles pose and appearance information, making independent control over them feasible. We now elaborate on the model design for these two steps.

{\bf Pose-guided canonical mapping.}
We define the human 2D image with SMPL pose $\mathbf{p}_o$ as the \textit{observation} space. To relieve learning difficulties, our model attempts to deform the observation space to a \textit{canonical} space with a predefined template pose $\mathbf{p}_c$ that is shared across different identities. The deformation function $T: \mathbb{R}^3 \mapsto \mathbb{R}^3$ thus maps spatial points $\mathbf{x}_i$ sampled in the observation space to $\mathbf{x}_i'$ in the canonical space. 

Learning such a deformation function has been proved effective for dynamic scene modeling~\cite{park2021nerfies,pumarola2021d}. However, learning to deform in such an implicit manner cannot  handle large articulation of humans and thus hardly generalizes to novel poses. To overcome this limitation, we use the SMPL   model to explicitly guide the deformation~\cite{liu2021neural,peng2021animatable,chen2021animatable}.
SMPL defines a skinned vertex-based human model   $(\mathcal{V},\mathcal{W})$, where $ \mathcal{V} = \{\mathbf{v} \} \in \mathbb{R}^{N \times 3}$ is the set of $N$  vertices  and $  \mathcal{W} = \{\mathbf{w}\} \in \mathbb{R}^{N \times K}$ is the set of the skinning weights  assigned for the vertex w.r.t.\ $K$ joints, with  $\sum_j w_j=1, w_j \geq 0$ for every joint.

We use the inverse-skinning (IS) transformation to map the SMPL mesh in the observation space with pose $\mathbf{p}$ into the canonical space~\cite{huang2020arch}:
\begin{equation}
    T_{\text{IS}} (\mathbf{v}, \mathbf{w}, \mathbf{p}) = \sum_j w_j \cdot (R_j\mathbf{v}+\mathbf{t}_j),
\end{equation}
where $R_j$ and $\mathbf{t}_j$ are the rotation and translation at each joint $j$ derived from  SMPL  with pose $\mathbf{p}$.

Such formulation can be easily extended to any spatial points in the observation space by simply adopting the same transformation from the nearest point on the surface of SMPL mesh. Formally,
for each spatial points $\mathbf{x}_i$, we first find its nearest point $\mathbf{v^{*}}$ on the SMPL mesh surface as $\mathbf{v^{*}}=\argmin_{\mathbf{v}\in \mathcal{V}}||\mathbf{x}_i-\mathbf{v}||_2$. Then, we use the corresponding skinning weights $\mathbf{w^*}$ to deform $\mathbf{x}_i$ to $\mathbf{x}_i'$ in the canonical space as:
\begin{equation}
    \mathbf{x}_i' = T_o (\mathbf{x}_i| \mathbf{p}) = T_{\text{IS}} ({\mathbf{x}_i}, \mathbf{w^*}, \mathbf{p}).
    \label{eqn:invskin}
\end{equation}

Although the SMPL-guided inverse-skinning transformation can help align the rigid skeleton with the template pose, 
it lacks the ability to model the pose-dependent deformation, like cloth wrinkles.
Besides, different identities may have different SMPL shape parameters $\mathbf{b}$, which likely leads to inaccurate transformation.

To alleviate these issues, \nameofmethod{} further trains a deformation network to model the residual deformation to complete the fine-grained geometric deformation and to compensate the inaccurate inverse-skinning transformation by
\begin{equation}
   \Delta \mathbf{x}_i' =  T_{\Delta}(\mathbf{x}_i'|\mathbf{w},\mathbf{p},\mathbf{b}) = \text{MLPs}(\text{Concat}[\text{Embed}(\mathbf{x}_i'),\mathbf{w},\mathbf{p},\mathbf{b}]),
\end{equation}
where $\mathbf{w}$ is the canonical style code mapped from the input latent code $\mathbf{z}$. We concatenate it with the embedded $\mathbf{x}_i'$ and  SMPL pose $\mathbf{p}$ and shape $\mathbf{b}$ parameters  and feed them to MLPs to yield the residual deformation. 
The final pose-guided deformation $T_{o\rightarrow c}$ from the observation  to canonical spaces can be formulated as
\begin{equation}
    T_{o\rightarrow c}(\mathbf{x}_i)=\mathbf{x}_i'+\Delta \mathbf{x}_i'=T_o (\mathbf{x}_i| \mathbf{p})+T_{\Delta}(T_o (\mathbf{x}_i|\mathbf{p})|\mathbf{w}, \mathbf{p},\mathbf{b}).
\end{equation}

{\bf Canonical avatar generation.}
After deforming 3D points sampled in the observation space to  the canonical space, we apply \nameofmethod{}   with the tri-plane representation for canonical avatar generation.
More concretely, it first generates tri-plane via a StyleGAN generator by taking the latent code $\mathbf{z}$ and camera parameters $\mathbf{c}$ as inputs. Then, for each point deformed via SMPL  parameters $\mathbf{p}$ in the canonical space, the model queries tri-plane to obtain the intermediate feature and maps it to color-based feature $c$ and density $\sigma$ for volume rendering. As such, it generates clothed human appearance and geometry in the canonical space with a predefined canonical pose, which alleviates the optimization difficulties and substantially helps our learning of high-quality avatar generation with disentangled pose and appearance control.

\subsection{Geometry-aware Human Modeling}
\label{sec:sdf}
To improve  geometry modeling of \nameofmethod{}, inspired by recent neural implicit surface works~\cite{wang2021neus,yariv2021volume,or2021stylesdf,peng2022animatable}, we adopt signed distance field (SDF) instead of density field as our geometry proxy, because it introduces more direct geometry regularization and guidance. To achieve this, our model learns to predict signed distance value rather than density in tri-plane for volume rendering.

\begin{figure*}[t]
    \centering
    \small
    \includegraphics[width=0.95\linewidth]{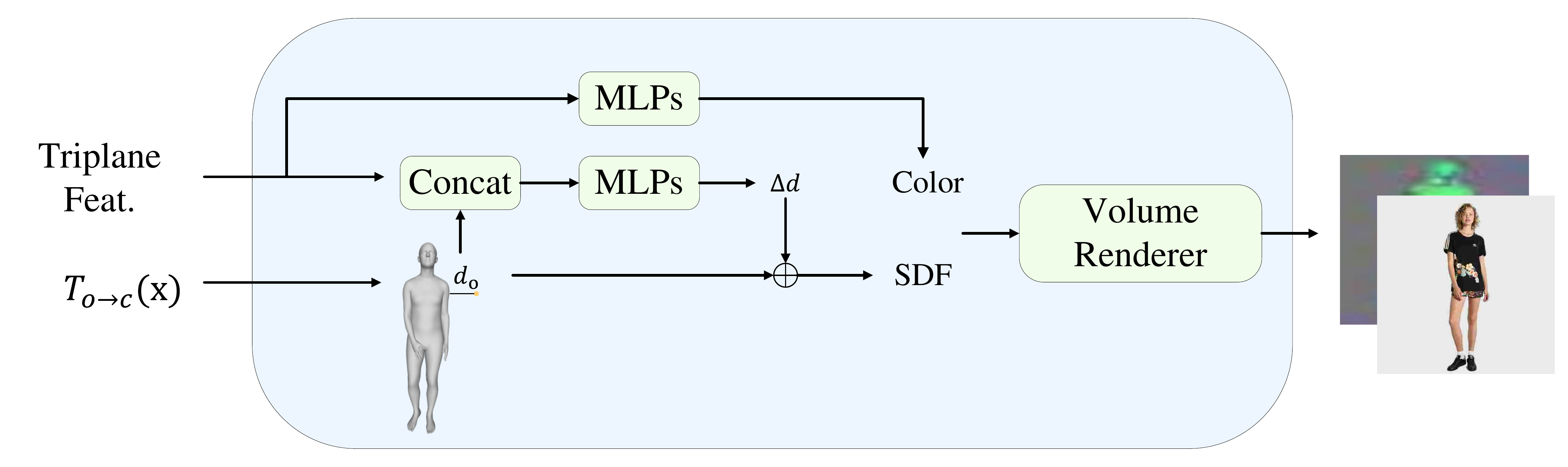}
    \caption{Our proposed geometry-aware human modeling module. It first predicts color and SDF values using the sampled tri-plane features and corresponding point location in the canonical space. Then it feeds them to volume renderer module to generate the raw image and features. The color for each sampled point is directly predicted from the tri-plane feature with MLPs. We use the SMPL model to guide the prediction of SDF and obtain a coarse signed distance value $d_{
   o}$ which is concatenated with the input tri-plane feature for predicting the residual distance $\Delta d$. The final SDF is   $d_
   o+\Delta d$. See \ref{sec:sdf} for more details.}
    \label{fig:sdf}
\end{figure*}

{\bf SMPL-guided geometry learning.}
Although SDF has a well-defined surface representation and introduces several regularization for geometry learning, how to use it for generative human modeling is still non-trivial due to  the complicated  body articulation and pose-dependent deformation. We therefore leverage the SMPL model as a guidance for the geometry-aware generation and combine it with a  residual SDF network (as shown in Fig.~\ref{fig:sdf}), that models the surface details (including hair and clothing)    not represented by SMPL. 

Specifically, given the input SMPL pose $\mathbf{p}_o$ and shape $\mathbf{b}_o$, we generate a SMPL mesh $M=T_\text{SMPL}(\mathbf{p}_o,\mathbf{b}_o)$, where $T_\text{SMPL}$ is the SMPL transformation function. For each 3D point $\mathbf{x}$ in the observation space, we first obtain its coarse signed distance value $d_o$ by querying the SMPL mesh $M$. Then, we feed $d_{o}$ alone with the   features from tri-plane to a light-weight MLP to predict the residual SDF $\Delta d$. The   signed distance value of each point is   computed as $d=d_{o}+\Delta d$. Predicting SDF with the coarse SMPL as guidance improves geometry learning of the model, thus achieving better human generation and animation, as demonstrated in our experiments.
We also introduce a SMPL-guided regularization for SDF learning as elaborated in Sec.~\ref{sec: training}.

{\bf SDF-based volume rendering.}
Following~\cite{or2021stylesdf}, we adopt SDF-based volume rendering to obtain the final output  images. 
For any point ${x}$ on the sampled rays, we  first deform it  to $\bar{x}=T_{o\rightarrow c}(x)$ by pose-guided canonical mapping. We query  feature vector $F(\bar{x})$ for position $\bar{x}$ from the canonical tri-plane and then feed it into two MLP layers to predict the color feature $c=\text{MLP}_c(F(\bar{x}))$ and the signed distance $d=d_o+\Delta d=d_o + \text{MLP}_{d}(F(\bar{x}), d_o)$. 
We then convert the signed distance value $d_i$ of each point $\mathbf{x}_i$ alone a ray $r$ to density value $\sigma_i$ as
$
\sigma_i = \frac{1}{\alpha}\cdot \text{Sigmoid}(\frac{-d_i}{\alpha}),
$
where $\alpha >0$ is a learnable parameter that controls the tightness of the density around the surface boundary.
By integration along the ray $r$ we can get the corresponding pixel feature as 
\begin{equation}
    I(r) = \sum_{i=1}^N \left(\prod_{j=1}^{i-1} e^{-\sigma_j \cdot \delta_j}\right) \cdot \left(1-e^{-\sigma_i \cdot \delta_i}\right) \cdot c_i,
\end{equation}
where $\delta_i = || \mathbf{x}_i-\mathbf{x}_{i-1} ||$. By aggregating all rays, we can get the entire image feature which is then feed into a StyleGAN decoder~\cite{Karras2020stylegan2} to generate the final high-resolutions synthesized image.

\subsection{Training}
\label{sec: training}
We use the non-saturating GAN loss $L_{\text{GAN}}$~\cite{Karras2020stylegan2} with R1 regularization $L_{\text{Reg}}$~\cite{goodfellow2014generative,mescheder2018training} to train our model  end-to-end. We also adopt the dual-discriminator proposed by EG3D~\cite{chan2021efficient}. It feeds both the rendered raw image and the decoded high-resolution image into the discriminator 
for improving   consistency of the generated  multi-view   images. 
To obtain better controllability, we feed both SMPL pose parameters $\mathbf{p}$ and camera parameters $\mathbf{c}$ as conditions to the discriminator for adversary training.
To regularize the learned SDF, we apply eikonal loss to the sampled points as:
\begin{equation}
    L_{\text{Eik}} = \sum_{\mathbf{x}_i} (||\nabla d_i|| -1)^2,
\end{equation}
where $\mathbf{x}_i$ and $d_i$ denote the sampled point  and predicted signed distance value, respectively. Following~\cite{or2021stylesdf}, we  adopt a minimal surface loss to encourage the model to represent human geometry with minimal volume of zero-crossings. It penalizes  the SDF value close to zero: 
\begin{equation}
L_{\text{Minsurf}} = \sum_{\mathbf{x}_i}\exp(-100 d_i).
\end{equation}
To make sure the generated surface is consistent with the input SMPL model, we incorporate the SMPL mesh as geometric prior and guide the generated surface to be close to the body surface.
Specifically, we sample vertices $\mathbf{v}\in \mathcal{V}$
on the SMPL body surface and then use it as query to deform to canonical space and sample features from the generated tri-plane and minimize the  signed distance. 
\begin{equation}
    L_{\text{SMPL}} = \sum_{\mathbf{v}\in \mathcal{V}}||\text{MLP}_d(F(T_{o\rightarrow c}(\mathbf{v}))) ||.
    \label{eqn:smpl}
\end{equation}

The overall loss is finally formulated as
\begin{equation}
    L_{\text{total}} = L_{\text{GAN}}+\lambda_{\text{Reg}}L_{\text{Reg}}+\lambda_{\text{Eik}}L_{\text{Eik}}+\lambda_{\text{Minsurf}}L_{\text{Minsurf}}+\lambda_{\text{SMPL}}L_{\text{SMPL}},
\end{equation}
where $\lambda_{*}$ are the corresponding loss weights.

\section{Experiments}
\label{experiments}

We study the following four questions in our experiments.
1) Is \nameofmethod{} able to generate 3D human avatars with realistic appearance and geometry?
2) Is \nameofmethod{} effective at controlling  human avatars  poses?
3) How does each component of our \nameofmethod{} model take effect?
4) Does \nameofmethod{} enable downstream applications, like single-view 3D reconstruction and text-guided synthesis? 
To answer these questions, we conduct extensive experiments on several 2D human fashion datasets~\cite{dong2019towards,liuLQWTcvpr16DeepFashion,zablotskaia2019dwnet}.

{\bf Datasets.} 
We evaluate  methods of 3D-aware clothed human generation on three real-world fashion datasets: MPV~\cite{dong2019towards}, DeepFashion~\cite{liuLQWTcvpr16DeepFashion} and UBCFashion~\cite{zablotskaia2019dwnet}. They contain single clothed people in each image.
We align and crop images according to the 2D human body keypoints, following~\cite{fu2022styleganhuman}. 
Since we focus on human avatar generation, we use a segmentation model~\cite{paddleseg2019} to remove irrelevant backgrounds.
We adopt an off-the-shelf pose estimator~\cite{kolotouros2019spin}  to obtain approximate camera and SMPL parameters.
We filter out images with partial observations and those with poor SMPL estimations, and get nearly 15K, 14K and 31K full-body images for each dataset, respectively.
Horizontal-flip augmentation is used during training.
We note these datasets are primarily composed of front-view images—few images captured from side or back views.
To compensate this, we sample more side- and back-view images to re-balance viewpoint distributions following~\cite{chan2021efficient}. 
We will release pre-processed scripts and datasets. 
For more details, please refer to the appendix.

\subsection{Comparisons}
\label{sec:mainresults}
{\bf Baselines.}
We compare our \nameofmethod{} against four state-of-the-art methods for 3D-aware image synthesis: EG3D~\cite{chan2021efficient}, StyleSDF~\cite{or2021stylesdf}, StyleNeRF~\cite{gu2021stylenerf} and GIRAFFE-HD~\cite{xue2022giraffe}. All these methods combine volume renderer with 2D decoder for 3D-aware image synthesis. EG3D and StyleNeRF adopt progressive training to improve performance. StyleSDF uses SDF as geometry representation for regularized geometry modeling.

\begin{table}[t]
    \centering
    \renewcommand{\tabcolsep}{3pt}
    \resizebox{1\textwidth}{!}{
    \begin{tabular}{@{\hskip 1mm}l c c c c| c c c c| c c c c@{\hskip 1mm}}
	\toprule
	& \multicolumn{4}{c|}{MPV} & \multicolumn{4}{c|}{DeepFashion} & \multicolumn{4}{c}{UBCFashion}\\
    & FID $\!\downarrow$          & Depth $\!\downarrow$ & Pose $\!\downarrow$ & Warp $\!\downarrow$   & FID $\!\downarrow$  & Depth $\!\downarrow$ & Pose $\!\downarrow$ & Warp  $\!\downarrow$ & FID $\!\downarrow$  & Depth $\!\downarrow$ & Pose $\!\downarrow$  & Warp $\!\downarrow$ \\
    \midrule
    GIRAFFE-HD & 26.3 & 2.12 & .099 & 31.4 & 25.3 & 1.94 & .092    & 34.3   & 27.0   & 2.03  & .094  & 35.2   \\
    StyleNeRF & 10.7 & 1.46 & .069 & 26.2 & 20.6 & 1.44 & .067 & 22.8 & 15.9 & 1.43 & .065 & 20.5   \\
    StyleSDF	            & 29.5 				    	& 1.74 	            & .648 					& 19.8 				    & 41.0     & 1.69 & .613    &  20.4  & 35.9 & 1.76 & .611 & 13.0   \\
    EG3D	            & 18.6 				    	& 1.52 	            & .077 					& 20.3 				    & 16.2     & 1.70 & .065    & 14.8   & 17.7  & 1.66 & .070  & 23.9   \\
    \midrule
    \nameofmethod{} (Ours) 			& \textbf{6.5} 				    	& \textbf{0.83} 	            & \textbf{.050} 					& \textbf{4.7} 				    & \textbf{9.6}     & \textbf{0.86} & \textbf{.052}   & \textbf{6.9}   & \textbf{8.7} & \textbf{0.94} & \textbf{.059} & \textbf{6.0}    \\
		\bottomrule
    \end{tabular}
    }
    \caption{Quantitative evaluation in terms of FID, depth, pose and warp accuracy on three datasets. Our \nameofmethod{} outperforms all the baselines significantly.
    }
    \label{tab:main_results}
\end{table}

{\bf Quantitative evaluations.}
Tab.~\ref{tab:main_results} provides quantitative comparisons between our \nameofmethod{} and the baselines. We measure image quality with Fr\'echet Inception Distance ({FID})~\cite{heusel2017gans} between 50k generated images and all of the available real images.  
We evaluate geometry quality by calculating Mean Squared Error  (MSE) against pseudo groundtruth (GT) depth-maps (\textit{Depth}) and poses (\textit{Pose}) that are estimated from the generated  images by~\cite{saito2020pifuhd,kolotouros2019spin}.
We also introduce an image warp metric (\textit{Warp}) that warps side-view image with depth map to frontal view and computes MSE against the generated frontal-view image to further evaluate the geometry quality and multi-view consistency of the model. For additional evaluation details, please refer to the appendix. 
From Tab.~\ref{tab:main_results}, we observe our model outperforms all the baselines w.r.t.\ all the metrics and   datasets. Notably, it outperforms baseline models by  significant margins (69.5\%, 63.1\%, 64.0\% in FID) on three datasets. These results clearly demonstrate its superiority in clothed human avatar synthesis. Moreover, it  maintains state-of-the-art geometry quality, pose accuracy and multi-view consistency.

{\bf Qualitative results.}
We show a qualitative comparison against baselines in the left of Fig.~\ref{fig:compare_inversion}. It can be observed that compared with our method, StyleSDF~\cite{or2021stylesdf} generate 3D avatar with over-smoothed geometry and poor multi-view consistency. In addition, the noise and holes can be observed around the generated avatar and the geometry details like face and clothes are missing.  EG3D~\cite{chan2021efficient} struggles to learn  3D human geometry from 2D images and suffers   degenerated qualities. 
Compared with them, our \nameofmethod{} generates 3D avatars with high-quality appearance with better view-consistency and geometric details.

\subsection{Ablation studies}
\label{sec:ablations}

We conduct ablation studies on the  Deepfashion dataset as its samples have  diverse poses and appearances.
We investigate effects of varying the following    designs   of  \nameofmethod{}.

\textbf{Geometry proxy.} 
Our \nameofmethod{} uses signed distance field (SDF) as geometry proxy to regularize the geometry learning. To investigate its effectiveness, we also evaluate our model with density field as the proxy. As shown in Tab.~\ref{ablation:geoproxy}, if replacing SDF with density field, the   quality of the generated avatars drops significantly\textemdash 11.1\% increase in FID, 38.6\% and 66.8\% increases in Depth and Warp metrics. This indicates SDF is important for the model to more precisely represent clothed human geometry. 
Without it, the model will produce noisy surface, and suffer  performance drop.

\textbf{Deformation schemes.} 
Our model uses a pose-guided deformation to transform spatial points from the observation  space to the canonical space. We also evaluate  other two deformation schemes in Tab.~\ref{ablation:deformation}: 1) residual deformation~\cite{park2021nerfies,park2020deformable} only (\textit{RD}), 2) inverse-skinning deformation~\cite{huang2020arch} only (\textit{IS}). 
When using RD only, the model training does not converge, indicating that learning deformation implicitly cannot handle large articulation of humans and   lead to implausible results.
While using IS only, the model achieves a reasonable result (FID: 10.7, Depth: 0.93, Warp: 7.7), verifying the importance of the explicitly pose-guided deformation. 
Further combining IS and RD  (our model) boosts the performance sharply\textemdash 10.3\%, 7.5\% and 10.4\% decrease in FID, Depth and Warp metrics, respectively. 
Introducing the residual deformation to collaborate with the posed-guided inverse-skinning transformation indeed  better represents non-rigid clothed human body deformation and thus our \nameofmethod{} achieves better appearance and geometry modeling.

\textbf{Number of KNN neighbors in inverse skinning deformation.}
For any spatial points, we use Nearest Neighbor to find the corresponding skinning weights for inverse skinning transformation (Eqn.~(\ref{eqn:invskin})). More nearest neighbors can be used for obtaining skinning weights~\cite{liu2021neural}. Thus, we study how the number of KNN affects model performance in Tab.~\ref{ablation:knn}. We observe using more KNN neighbors gives worse performance. This is likely caused by 1) noisy skinning weights introduced by using more neighbors for calculation and 2) inaccurate SMPL estimation in data pre-processing step. 

\begin{table}[t]
	\renewcommand{\tabcolsep}{2pt}
	\small
	\begin{subtable}[!t]{0.29\textwidth}
		\centering
		\begin{tabular}{cccccccc}
			\toprule
			\textit{Geo.} & FID & Depth & Warp \\
			\midrule
			Density  & 10.8 & 1.40 & 20.8\\
			SDF  & 9.6 & 0.86 & 6.9\\
			\bottomrule
		\end{tabular}
		\caption{The effect of different geometry proxies.
		}
		\label{ablation:geoproxy}
	\end{subtable}
	\hspace{\fill}
	\begin{subtable}[!t]{0.373\textwidth}
		\centering
		\begin{tabular}{cccccccc}
			\toprule
			\textit{Deform.} & FID & Depth & Warp \\
			\midrule
			RD & - & - & -\\
			IS & 10.7 & 0.93 & 7.7\\
			IS+RD & 9.6 & 0.86 & 6.9\\
			\bottomrule
		\end{tabular}
		\caption{Deformation schemes. IS and RD are inverse skinning and residual deformation.
		}\vspace{-10pt}
		\label{ablation:deformation}
	\end{subtable}
	\hspace{\fill} 
	\begin{subtable}[!t]{0.3\textwidth}
		\centering
		\begin{tabular}{ccccccccc}
			\toprule
			\textit{KNN} & FID & Depth & Warp  \\
			\midrule
			1 & 9.6 & 0.86 & 6.9\\
			2 & 10.4 & 0.90 & 7.4 \\
			3 & 13.1 & 1.08 & 10.2 \\
			4 & 16.3 & 1.14 & 15.3 \\
			\bottomrule
		\end{tabular}
		\caption{Different number of KNN in inverse skinning deformation. 
		}\vspace{-20pt}
		\label{ablation:knn}
	\end{subtable}
	\hspace{\fill}
	\begin{subtable}[b]{0.29\textwidth}
		\centering
		\begin{tabular}{cccccccc}
			\toprule
			\textit{Ray Steps} & FID & Depth & Warp \\
			\midrule
			12 & 12.4 & 1.04 & 7.7\\
			24 & 10.7 & 0.92 & 7.5\\
			36 & 10.0 & 0.89 & 7.2\\
			48 & 9.6 & 0.86 & 6.9\\
			\bottomrule
		\end{tabular}
		\caption{Number of ray steps. 
		}
		\label{ablation:ray_step}
	\end{subtable}
	\hspace{\fill}
	\begin{subtable}[b]{0.373\textwidth}
		\centering
		\begin{tabular}{cccccccc}
			\toprule
			\textit{SDF Prior} & FID & Depth & Warp \\
			\midrule
			w/o & 14.3 & 1.12 & 8.3\\
			Can. & 10.4 & 0.89 & 7.5 \\
			Obs. & 9.6 & 0.86 & 6.9\\
			\bottomrule
		\end{tabular}
		\caption{The effect of SMPL body SDF priors. 
		}
		\label{ablation:bodysdf}
	\end{subtable}
	\hspace{\fill}
	\begin{subtable}[b]{0.3\textwidth}
		\centering
		\begin{tabular}{cccccccc}
			\toprule
			\textit{SDF Scheme} & FID & Depth & Warp \\
			\midrule
			Raw & 10.8 & 0.94 & 7.9\\
			Residual & 9.6 & 0.86 & 6.9\\
			\bottomrule
		\end{tabular}
		\caption{SDF prediction schemes. 
		}
		\label{ablation:sdfres}
	\end{subtable}
	\hspace{\fill}
	\caption{Ablations on Deepfashion. In (e), \textit{w/o} denotes without using SMPL body SDF prior, \textit{Can.} or \textit{Obs.} means using  SDF prior from canonical or observation spaces. 
	}
	\label{ablations}
	\vspace{-8mm}
\end{table}

\textbf{Number of ray steps.}  Tab.~\ref{ablation:ray_step} shows the effect of the number of points sampled per ray for volume rendering. With only 12 sampled points for each ray, \nameofmethod{} already achieves acceptable results, \ie, 12.4, 1.04 and 7.7 in FID, Depth and Warp losses. With more sampling points, the performance monotonically increases, demonstrating  the  capacity of \nameofmethod{} in 3D-aware avatars generation.

\textbf{SMPL body SDF priors.}  
\nameofmethod{} adopts a SMPL-guided geometry learning scheme, \ie, generating clothed human body SDFs on top of the coarse SMPL body mesh. As shown in Table~\ref{ablation:bodysdf}, if removing SMPL body guidance, the performance drops significantly, \ie, 32.9\%, 23.2\%, 16.9\% increase in FID, Depth and Warp losses. This indicates the coarse SMPL body information is   important for guiding \nameofmethod{} to better generate clothed human geometry. 
We also evaluate the performance difference between SMPL body SDFs queried from observation (\textit{Obs.}) or canonical (\textit{Can.}) spaces. We see the model guided by body SDFs queried from observation space obtains better performance as they are more accurate than the ones queried from the canonical space.
Moreover, we study the effect of SMPL SDF regularization loss in Eqn.~(\ref{eqn:smpl}). If removing the regularization loss, the performance in all metrics drops (FID: 10.8 \emph{vs.} 9.6, Depth: 0.96 \emph{vs.} 0.86, Warp: 7.8 \emph{vs.} 6.9), verifying the effectiveness of the proposed loss for regularized geometry learning.

\textbf{SDF prediction schemes.}  Table~\ref{ablation:sdfres} shows the effect of two SDF prediction schemes\textemdash predicting raw SDFs directly or SDF residuals on top of the coarse SMPL body SDFs. Compared with predicting the raw SDFs directly, the residual prediction scheme delivers better results, since it alleviates the geometry learning difficulties.

\subsection{Applications}
\label{sec:applications}
\begin{figure*}[t]
    \centering
    \small
    \includegraphics[width=0.95\linewidth]{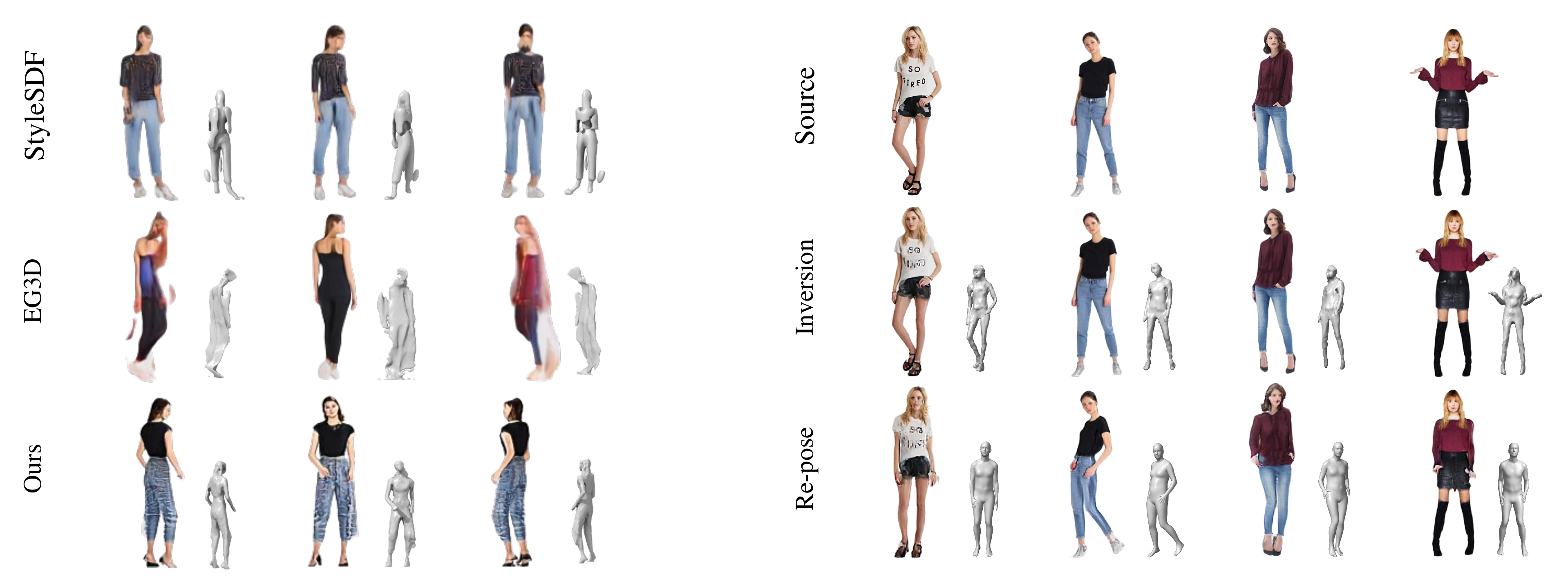}
 \caption{(Left) Qualitative comparison of multi-view rendering and geometry quality against baselines including EG3D~\cite{chan2021efficient} and StyleSDF~\cite{or2021stylesdf}. (Right) Single-view 3D  reconstruction and reanimation result of \nameofmethod{}. Given source image, we reconstruct both color and geometry of the human in the image. The re-pose step further takes novel SMPL parameters as input and animates the reconstructed avatar. 
 } 
    \label{fig:compare_inversion}
\end{figure*}

{\bf Single-view 3D reconstruction and re-pose.}
The right panel of Fig.~\ref{fig:compare_inversion} shows the application of our learned latent space for single-view 3D reconstruction. Following~\cite{chan2021efficient}, we use pivotal tuning inversion (PTI)~\cite{roich2021pivotal} to fit the target images (top) and recover both the appearance and the geometry (middle). With the recovered 3D representation and latent code, we can further use novel SMPL parameters (bottom) to re-pose/animate the human in the source images.

\begin{figure*}[h]
    \centering
    \small
    \includegraphics[width=0.95\linewidth]{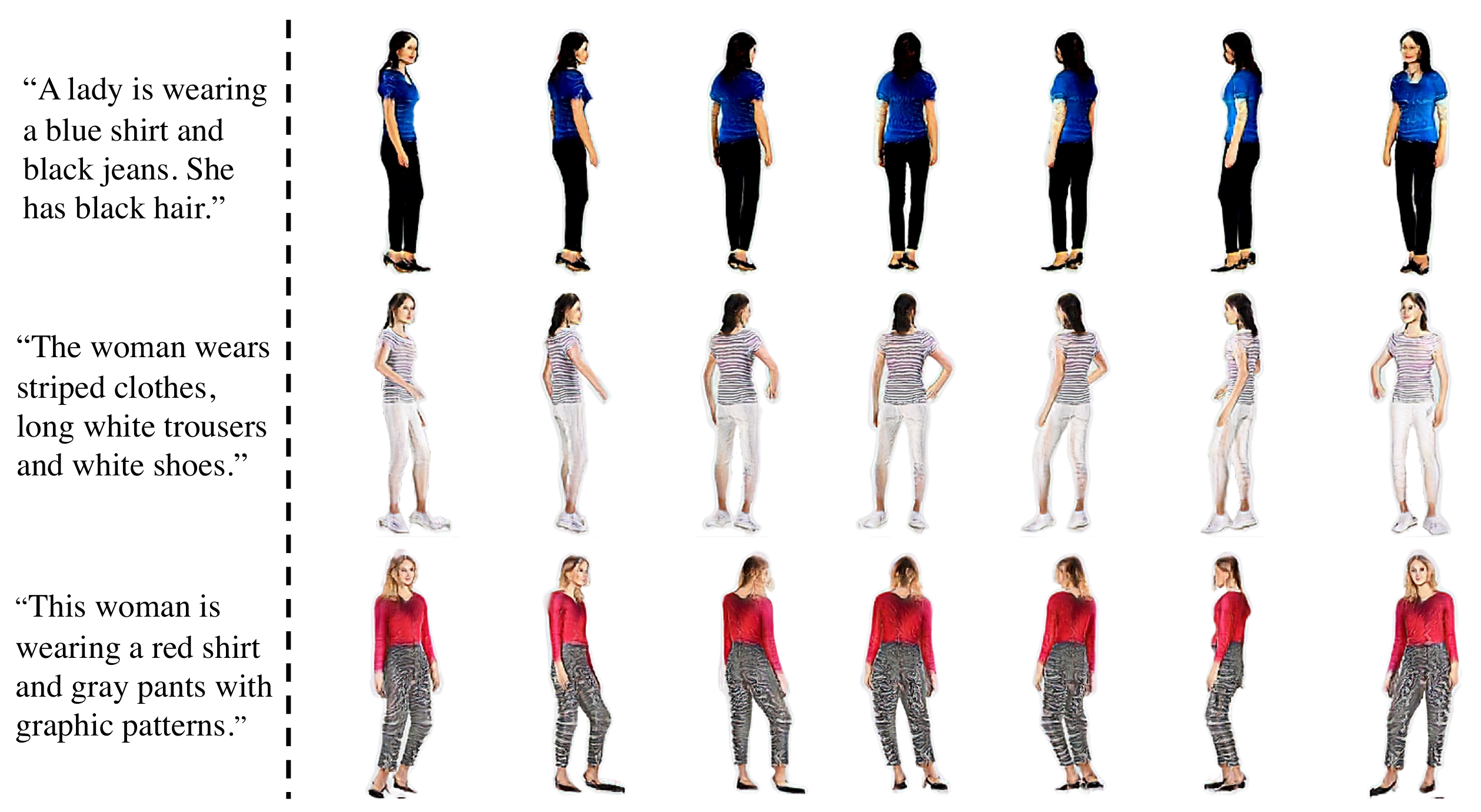}
 \caption{Text-guided (left) synthesis results of \nameofmethod{} with multi-view rendering (right).} 
    \label{fig:clip_inversion}
\end{figure*}

{\bf Text-guided synthesis.}
Recent works~\cite{Patashnik_2021_ICCV,kwon2022clipstyler} have shown that one could use a text-image embedding, such as CLIP~\cite{radford2021learning}, to guide StyleGAN2 for controlled synthesis. We also visualize text-guided clothed human synthesis in Fig.~\ref{fig:clip_inversion}.  Specifically, we use StyleCLIP~\cite{Patashnik_2021_ICCV} to manipulate a synthesized image with a sequence of text prompts. The optimization based StyleCLIP is used as it is flexible for any input text. From the figure, our \nameofmethod{} is able to synthesis different style human images given different text prompts. This clearly indicates that  
\nameofmethod{} can be an effective  tool for text-guided portrait synthesis where detailed descriptions are provided.

\section{Conclusion}
\label{sec:conclusion}

This work introduced the first 3D-aware human avatar generative model, \nameofmethod{}. By factorizing the generative process into the canonical avatar generation and deformation stages, \nameofmethod{}  can leverage the  geometry prior and  effective tri-plane representation to  address  the challenges in animatable human avatar generation.  We demonstrated \nameofmethod{} can generate clothed human avatars with arbitrary poses and viewpoints. Besides, it can also generate avatars from multi-modality input conditions, like natural language description and 2D images (for inverting). This work substantially extends the 3D generative models from objects of simple structures (e.g., human faces, rigid objects) to articulated and complex objects. We believe this model will make the creation of human avatars more accessible to ordinary users, assist designers  and reduce the manual cost.

{\small
\bibliographystyle{plain}
\bibliography{reference}
}

\end{document}